\documentclass[11pt,a4paper]{article}
\usepackage[hyperref]{emnlp-ijcnlp-2019}
\usepackage{times}
\usepackage{latexsym}

\usepackage{url}

\usepackage{helvet}
\usepackage{courier}

\renewcommand{\vec}[1]{\mathbf{#1}}

\usepackage{multirow}
\usepackage{hhline}
\usepackage{tabularx}
\usepackage{ragged2e}
\newcolumntype{Y}{>{\RaggedRight\let\newline\\\arraybackslash\hspace{0pt}}X} 
\usepackage{booktabs}
\usepackage{tkz-berge}
\usepackage{xcolor}
\usepackage{pgfplots}
\usepackage{pgfpages}
\usepackage{subcaption}
\usepackage{mathbbol}
\usepackage{CJKutf8}
\usepackage{arydshln}
\usepackage{amsmath}

\usepackage{amssymb}
\usepackage{amsfonts}

\usepackage{algorithm}
\usepackage{algorithmic}
\usepackage{tkz-berge}
\usepackage{tikz}
\usepackage{pgf}
\usepackage{tikz-qtree}
\usepackage{xcolor}
\usetikzlibrary{arrows,decorations.pathmorphing,backgrounds,positioning,fit,petri,shapes.misc, arrows.meta,shapes.geometric,decorations.markings,calc,shadows.blur,decorations.pathreplacing,quotes,matrix,shapes.symbols}
\definecolor{fontgray}{RGB}{44, 62, 80}
\definecolor{myred}{RGB}{235, 47, 6} 
\definecolor{summertime}{RGB}{245, 205, 121}
\definecolor{darkgrass}{RGB}{0, 148, 50}
\definecolor{myblue}{RGB}{0, 168, 255}
\definecolor{mygray}{RGB}{158, 158, 158}
\definecolor{puffin}{RGB}{250, 152, 58}
\definecolor{lowpurple}{RGB}{210, 180, 222}
\definecolor{lowblue}{RGB}{102,178,255}
\definecolor{lowred}{RGB}{245, 183, 177}

\usepackage{lipsum}
\usepackage{multicol}
\usepackage{changepage}

\usepackage[export]{adjustbox}
\usepackage{booktabs, tabularx}

\usepackage[utf8]{inputenc}
\usepackage[english]{babel}
 
\usepackage{amsthm}

\newcommand{\pluseq}{\mathrel{+}=}

\newcounter{Lcount}
\newcommand{\squishenum}{
\begin{list}{\arabic{Lcount}. }
{ \usecounter{Lcount}
\setlength{\itemsep}{0pt}
\setlength{\parsep}{0pt}
\setlength{\topsep}{0pt}
\setlength{\partopsep}{0pt}
\setlength{\leftmargin}{2em}
\setlength{\labelwidth}{1.5em}
\setlength{\labelsep}{0.5em} } }

\newcommand{\RNum}[1]{\uppercase\expandafter{\romannumeral #1\relax}}
\newcommand{\squishletter}{
\begin{list}{\alph{Lcount}. }
{ \usecounter{Lcount}
\setlength{\itemsep}{0pt}
\setlength{\parsep}{0pt}
\setlength{\topsep}{0pt}
\setlength{\partopsep}{0pt}
\setlength{\leftmargin}{2em}
\setlength{\labelwidth}{1.5em}
\setlength{\labelsep}{0.5em} } }

\newcommand{\squishlist}{
\begin{list}{$\bullet$}
{ \usecounter{Lcount}
\setlength{\itemsep}{0pt}
\setlength{\parsep}{0pt}
\setlength{\topsep}{0pt}
\setlength{\partopsep}{0pt}
\setlength{\leftmargin}{2em}
\setlength{\labelwidth}{1.5em}
\setlength{\labelsep}{0.5em} } }

\DeclareMathOperator*{\softmax}{softmax}

\newcommand{\squishend}{
\end{list} }

\makeatletter
\def\blfootnote{\xdef\@thefnmark{}\@footnotetext}
\makeatother

\aclfinalcopy 

\title{Learning Explicit and Implicit Structures for Targeted Sentiment Analysis}

\author{Hao Li \and Wei Lu \\
  StatNLP Research Group \\
  Singapore University of Technology and Design \\
  {{\tt hao\_li@mymail.sutd.edu.sg}}\\{ {\tt luwei@sutd.edu.sg}} \\}

\date{}

\begin{document}
\maketitle
\begin{abstract}
Targeted sentiment analysis is the task of jointly predicting target entities and their associated sentiment information.
Existing research efforts mostly regard this joint task as a sequence labeling problem, building models that can capture explicit structures in the output space.
However, the importance of capturing implicit global structural information that resides in the input space is largely unexplored.
In this work, we argue that both types of information (implicit and explicit structural information) are crucial for building a successful targeted sentiment analysis model.
Our experimental results show that properly capturing both information is able to lead to better performance than competitive existing approaches.
We also conduct extensive experiments to investigate our model's effectiveness and robustness\footnote{We release our code at \url{http://www.statnlp.org/research/st}.}.
\end{abstract}

\section{Introduction}
\blfootnote{Accepted as a long paper in EMNLP 2019 (Conference on Empirical Methods in Natural Language Processing).}

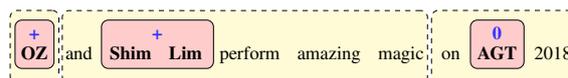
\begin{figure}[t!]
		\centering
		\adjustbox{max width=1.0\linewidth}{
			\begin{tikzpicture}[node distance=1.0mm and 1.0mm, >=Stealth, 
			wordnode/.style={draw=none, minimum height=5mm, inner sep=0pt},
			chainLine/.style={line width=1pt,-, color=fontgray},
			scopebox/.style={draw=black, rounded corners, fill=yellow!20, dashed},
			targetbox/.style={draw=black, rounded corners, fill=red!20}
			]

			\matrix (sent1) [matrix of nodes, nodes in empty cells, execute at empty cell=\node{\strut};]
			{
				\textbf{OZ} & [1mm] and &[1mm] \textbf{Shim} & [1mm] \textbf{Lim} &  [1mm]  perform & [1mm] amazing & [1mm] magic & [1mm]on & [1mm] \textbf{AGT} & [1mm]2018\\
			};
						
			\begin{pgfonlayer}{background}
			\node [scopebox, above=of sent1-1-1, yshift=-10mm, text height=-5mm, minimum height = 15mm, minimum width=10mm] (s1)  [] {\color{blue!80}\textbf{\textsc{}}};
			\node [targetbox, above=of sent1-1-1, yshift=-7mm, text height=-2mm, minimum height = 10mm, minimum width=8mm] (e1)  [] {\color{blue!80}\textbf{\textsc{+}}};

			\node [scopebox, above=of sent1-1-5, xshift=-1mm, yshift=-10mm, text height=-5mm, minimum height = 15mm, minimum width=75mm] (s2)  [] {\color{blue!80}\textbf{\textsc{}}};
			\node [targetbox, above=of sent1-1-3, xshift=5.5mm, yshift=-7mm, text height=-2mm, minimum height = 10mm, minimum width=23mm] (e2)  [] {\color{blue!80}\textbf{\textsc{+}}};

			\node [scopebox, above=of sent1-1-9, xshift=1.5mm, yshift=-10mm, text height=-5mm, minimum height = 15mm, minimum width=30mm] (s3)  [] {\color{blue!80}\textbf{\textsc{0}}};
			\node [targetbox, above=of sent1-1-9, xshift=0mm, yshift=-7mm, text height=-2mm, minimum height = 10mm, minimum width=11mm] (e3)  [] {\color{blue!80}\textbf{\textsc{0}}};
			
			\end{pgfonlayer}
			
			\end{tikzpicture} 
		}
		\caption{TSA with targets in bold and their associated sentiment on top. Boundaries for the sentiment scope are highlighted in dashed boxes.}
		\label{fig:example}
	\end{figure}

Targeted sentiment analysis (TSA) is an important task useful for public opinion mining~\cite{pang2008opinion,liu2010sentiment,ortigosa2014sentiment,smailovic2013predictive,li2010using}. 
The task focuses on predicting the sentiment information towards a specific target phrase, which is usually a named entity, in a given input sentence.
Currently, TSA in the literature may refer to either of the two possible tasks under two different setups: 1) predicting the sentiment polarity for a given specific target {\color{black}phrase~\cite{dong2014adaptive,wang2016recursive,zhang2016gated,xue2018aspect}}; 2) jointly predicting the targets together with the sentiment polarity assigned to each {\color{black}target~\cite{mitchell2013open,zhang2015neural,li2017sentimentscope,ma2018joint}}.
In this paper, we focus on the latter setup which was originally proposed by~\citet{mitchell2013open}. 
Figure~\ref{fig:example} presents an example sentence containing three targets.
Each target is associated with a sentiment, where we use $+$ for denoting positive polarity, $0$ for neutral and $-$ for negative.

Existing research efforts mostly regard this task as a sequence labeling problem by assigning a tag to each word token, where the tags are typically designed in a way that capture both the target boundary as well as the targeted sentiment polarity information together.
Existing approaches~\cite{mitchell2013open,zhang2015neural,ma2018joint} build models based on conditional random fields (CRF)~\cite{lafferty2001conditional} or structural support vector machines (SSVM)~\cite{taskar2005learning,tsochantaridis2005large} to explicitly model the sentiment information with structured outputs, where each targeted sentiment prediction corresponds to exactly one \textit{fixed} output. While effective, such models suffer from their inability in capturing certain long-distance dependencies between sentiment keywords and their targets. 
To remedy this issue,~\citet{li2017sentimentscope} proposed their ``sentiment scope’’ model to learn flexible output representations.
For example, three text spans with their corresponding targets in bold are presented in Figure~\ref{fig:example}, where each target’s sentiment is characterized by the words appearing in the corresponding text span. They learn from data for each target a latent text span used for attributing its sentiment, resulting in \textit{flexible} output structures.

However, we note there are two major limitations with the approach of~\citet{li2017sentimentscope}.
First, their model requires a large number of hand-crafted discrete features. 
Second, the model relies on a strong assumption that the latent sentiment spans do not overlap with one another.
For example, in Figure~\ref{fig:example}, their model will not be able to capture the interaction between the target word ``\textit{OZ}'' in the first sentiment span and the keyword ``\textit{amazing}'' due to the assumptions made on the explicit structures in the output space. 
One idea to resolve this issue is to design an alternative mechanism to capture such useful structural information that resides in the input space.

On the other hand, recent literature shows that feature learning mechanisms such as self-attention have been successful for the task of sentiment prediction when targets are given~\cite{wang2018learning,he-etal-2018-effective,fan-etal-2018-multi} (i.e., under the first setup mentioned above).
Such approaches essentially attempt to learn rich \textit{implicit} structural information in the input space that captures the interactions between a given target and all other word tokens within the sentence.
Such implicit structures are then used to generate sentiment summary representation towards the given target, leading to the performance boost.

However, to date capturing rich implicit structures in the joint prediction task that we focus on (i.e., the second setup) remains largely unexplored.
Unlike the first setup, in our setup the targets are not given, we need to handle exponentially many possible combinations of targets in the joint task.
This makes the design of an algorithm for capturing both implicit structural information from the input space and the explicit structural information from the output space challenging.

Motivated by the limitations and challenges, we present a novel approach that is able to efficiently and effectively capture the explicit and implicit structural information for TSA.
We make the following key contributions in this work:
\squishlist
   \item We propose a model that is able to properly integrate both \textit{explicit} and \textit{implicit} structural information, called \textbf{EI}. The model is able to learn flexible explicit structural information in the output space while being able to efficiently learn rich implicit structures by LSTM and self-attention for exponentially many possible combinations of targets in a given sentence.
    \item We conducted extensive experiments to validate our claim that both explicit and implicit structures are indispensable in such a task, and demonstrate the effectiveness and robustness of our model.

\squishend


\section{Approach}
Our objective is to design a model to extract targets as well as their associated targeted sentiments for a given sentence in a joint manner. 
As we mentioned before, we believe that both explicit and implicit structures are crucial for building a successful model for TSA.
Specifically, we first present an approach to learn flexible explicit structures based on latent CRF, and next present an approach to efficiently learn the rich implicit structures for exponentially many possible combinations of targets.

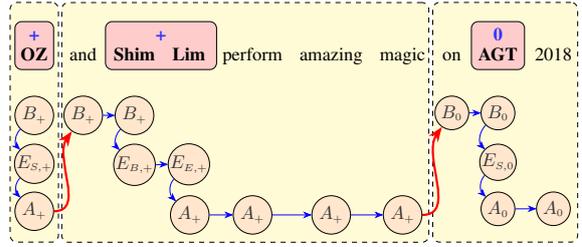
\begin{figure}[t!]
		\centering
		\adjustbox{max width=1.0\linewidth}{
			\begin{tikzpicture}[node distance=1.0mm and 1.0mm, >=Stealth, 
			wordnode/.style={draw=none, minimum height=5mm, inner sep=0pt},
			chainLine/.style={line width=1pt,-, color=fontgray},
			scopebox/.style={draw=black, rounded corners, fill=yellow!20, dashed, minimum height = 50mm},
			targetbox/.style={draw=black, rounded corners, fill=red!20, minimum height = 10mm},
			btag/.style={shape=circle, draw=black, rounded corners, fill=orange!20, minimum height=8mm, inner sep=0pt},
			atag/.style={shape=circle, draw=black, rounded corners, fill=orange!20, minimum height=8mm, inner sep=0pt},
			etag/.style={shape=circle, draw=black, rounded corners, fill=orange!20, minimum height=8mm, inner sep=0pt}
			]

			\matrix (sent1) [matrix of nodes, nodes in empty cells, execute at empty cell=\node{\strut};]
			{
				\textbf{OZ} & [2mm] and &[1mm] \textbf{Shim} & [1mm] \textbf{Lim} &  [1mm]  perform & [1mm] amazing & [1mm] magic & [1mm]on & [1mm] \textbf{AGT} & [1mm]2018\\
			};
			
			
			\foreach \pos in {1,2,3}
				\node [btag, below=of sent1-1-\pos, yshift=-5mm] (btag\pos)  [] {\color{black!80}\textsc{$B_{\scriptscriptstyle +}$}};
			\foreach \pos in {8,9}
				\node [btag, below=of sent1-1-\pos, yshift=-5mm, minimum height=7mm] (btag\pos)  [] {\color{black!80}\textsc{$B_{\scriptscriptstyle 0}$}};
			
			\foreach \pos in {1,5,6,7}
				\node [atag, below=of sent1-1-\pos, yshift=-25mm] (atag\pos)  [] {\color{black!80}\textsc{$A_{\scriptscriptstyle +}$}};
			\node [atag, below=of sent1-1-5, xshift=-13mm, yshift=-25mm] (atag4)  [] {\color{black!80}\textsc{$A_{\scriptscriptstyle +}$}};
			\foreach \pos in {9,10}
				\node [atag, below=of sent1-1-\pos, yshift=-25mm, minimum height=7mm] (atag\pos)  [] {\color{black!80}\textsc{$A_{\scriptscriptstyle 0}$}};

			\node [etag, below=of sent1-1-1, yshift=-15mm, minimum height=7mm] (etag1)  [] {\color{black!80}\textsc{$E_{\scriptscriptstyle S,+}$}};
			\node [etag, below=of sent1-1-3, yshift=-15mm, minimum height=7mm] (etag3)  [] {\color{black!80}\textsc{$E_{\scriptscriptstyle B,+}$}};
			\node [etag, below=of sent1-1-4, yshift=-15mm, minimum height=7mm] (etag4)  [] {\color{black!80}\textsc{$E_{\scriptscriptstyle E,+}$}};
			\node [etag, below=of sent1-1-9, yshift=-15mm, minimum height=7mm] (etag9)  [] {\color{black!80}\textsc{$E_{\scriptscriptstyle S,0}$}};

			
			\foreach \from/\to in {{btag1/etag1},{etag1/atag1},{btag3/etag3},{etag4/atag4},{btag9/etag9},{etag9/atag9}}
			\draw[->,blue]   (\from) to[out=225,in=135] (\to);
			
			\foreach \from/\to in {{btag2/btag3},{etag3/etag4},{atag4/atag5},{atag5/atag6},{atag6/atag7},{btag8/btag9},{atag9/atag10}}
			\draw[->,blue]   (\from) to[out=0,in=180] (\to);
			
			\foreach \from/\to in {{atag1/btag2},{atag7/btag8}}
			\draw[->,red,line width=1.3pt]   (\from) to[out=0,in=235] (\to);

			\begin{pgfonlayer}{background}
			\node [scopebox, above=of sent1-1-1, xshift=0mm, yshift=-43mm, text height=5mm, minimum width=10mm] (s1)  [] {\color{blue!80}\textbf{\textsc{}}};
			\node [targetbox, above=of sent1-1-1, xshift=0mm, yshift=-7mm, text height=-2mm, minimum width=8mm] (e1)  [] {\color{blue!80}\textbf{\textsc{+}}};

			\node [scopebox, above=of sent1-1-5, xshift=-1.5mm, yshift=-43mm, text height=-5mm, minimum width=76mm] (s2)  [] {\color{blue!80}\textbf{\textsc{}}};
			\node [targetbox, above=of sent1-1-3, xshift=5.5mm, yshift=-7mm, text height=-2mm, minimum width=23mm] (e2)  [] {\color{blue!80}\textbf{\textsc{+}}};

			\node [scopebox, above=of sent1-1-9, xshift=1.5mm, yshift=-43mm, text height=-5mm, minimum width=30mm] (s3)  [] {\color{blue!80}\textbf{\textsc{}}};
			\node [targetbox, above=of sent1-1-9, xshift=0mm, yshift=-7mm, text height=-2mm, minimum width=11mm] (e3)  [] {\color{blue!80}\textbf{\textsc{0}}};
			
			\end{pgfonlayer}

			\end{tikzpicture} 
		}
		\caption{The structured output for representing entities and their sentiments with boundaries.}
		\label{fig:ss_seq1}
	\end{figure}

\subsection{Explicit Structure}


Motivated by \citet{li2017sentimentscope}, we design an approach based on latent CRF {\color{black}to model flexible sentiment spans to capture} better explicit structures in the output space. 
To do so, we firstly integrate target and targeted sentiment information into a label sequence by using 3 types of tags in our \textbf{EI} model: $\mathbf{B}_p$, $\mathbf{A}_p$, and $\mathbf{E}_{\epsilon,p}$, where $p \in \{+, -, 0\}$ indicates the sentiment polarity and $\epsilon \in \{\textit{B,M,E,S}\}$ denotes the \textit{BMES} tagging scheme\footnote{\textit{B} stands for the beginning of the target phrase, \textit{M} for the middle, \textit{E} for the end and \textit{S} for a single-word target.}. We explain the meaning of each type of tags as follows.
\begin{itemize}
    \item $\mathbf{B}_p$ is used to denote that the current word is part of a sentiment span with polarity $p$, but appears before the target word or exactly as the first word of the target. 
    \item $\mathbf{A}_p$ is used to denote that the current word is part of a sentiment span with polarity $p$, but appears after the target word or exactly as the last word of the target. 
    \item $\mathbf{E}_{\epsilon,p}$ is used to denote the current word is part of a sentiment span with polarity $p$, and is also a part of the target. 
    The \textit{BMES} sub-tag $\epsilon$ denotes the position information within the target phrase. 
    For example, $\mathbf{E}_{B,+}$ represents that the current word appears as the first word of a target with the positive polarity.
\end{itemize}





We illustrate how to construct the label sequence for a specific combination of sentiment spans of the given example sentence in Figure~\ref{fig:ss_seq1}, where three non-overlapping sentiment spans in yellow are presented. 
Each such sentiment span encodes the sentiment polarity in blue for a target in bold in pink square. 
At each position, we allow multiple tags in a sequence to appear such that the edge $\mathbf{A}_p\mathbf{B}_{p'}$ in red consistently indicates the boundary between two adjacent sentiment spans. 

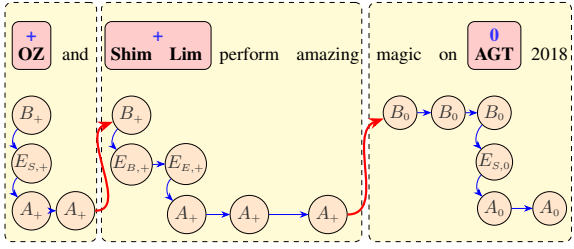
\begin{figure}[t!]
		\centering
		\adjustbox{max width=1.0\linewidth}{
			\begin{tikzpicture}[node distance=1.0mm and 1.0mm, >=Stealth, 
			wordnode/.style={draw=none, minimum height=5mm, inner sep=0pt},
			chainLine/.style={line width=1pt,-, color=fontgray},
			scopebox/.style={draw=black, rounded corners, fill=yellow!20, dashed, minimum height = 50mm},
			targetbox/.style={draw=black, rounded corners, fill=red!20, minimum height = 10mm},
			btag/.style={shape=circle, draw=black, rounded corners, fill=orange!20, minimum height=8mm, inner sep=0pt},
			atag/.style={shape=circle, draw=black, rounded corners, fill=orange!20, minimum height=8mm, inner sep=0pt},
			etag/.style={shape=circle, draw=black, rounded corners, fill=orange!20, minimum height=8mm, inner sep=0pt}
			]

			\matrix (sent1) [matrix of nodes, nodes in empty cells, execute at empty cell=\node{\strut};]
			{
				\textbf{OZ} & [1mm] and &[2mm] \textbf{Shim} & [1mm] \textbf{Lim} &  [1mm]  perform & [1mm] amazing & [1mm] magic & [1mm]on & [1mm] \textbf{AGT} & [1mm]2018\\
			};
			
			
			\foreach \pos in {1,3}
				\node [btag, below=of sent1-1-\pos, yshift=-5mm] (btag\pos)  [] {\color{black!80}\textsc{$B_{\scriptscriptstyle +}$}};
			\foreach \pos in {8,9}
				\node [btag, below=of sent1-1-\pos, yshift=-5mm, minimum height=7mm] (btag\pos)  [] {\color{black!80}\textsc{$B_{\scriptscriptstyle 0}$}};
			\node [btag, below=of sent1-1-8, xshift=-10mm, yshift=-5mm, minimum height=7mm] (btag7)  [] {\color{black!80}\textsc{$B_{\scriptscriptstyle 0}$}};
			
			\foreach \pos in {1,2,5,6}
				\node [atag, below=of sent1-1-\pos, yshift=-25mm] (atag\pos)  [] {\color{black!80}\textsc{$A_{\scriptscriptstyle +}$}};
			\node [atag, below=of sent1-1-5, xshift=-13mm, yshift=-25mm] (atag4)  [] {\color{black!80}\textsc{$A_{\scriptscriptstyle +}$}};
			\foreach \pos in {9,10}
				\node [atag, below=of sent1-1-\pos, yshift=-25mm, minimum height=7mm] (atag\pos)  [] {\color{black!80}\textsc{$A_{\scriptscriptstyle 0}$}};

			\node [etag, below=of sent1-1-1, yshift=-15mm, minimum height=7mm] (etag1)  [] {\color{black!80}\textsc{$E_{\scriptscriptstyle S,+}$}};
			\node [etag, below=of sent1-1-3, yshift=-15mm, minimum height=7mm] (etag3)  [] {\color{black!80}\textsc{$E_{\scriptscriptstyle B,+}$}};
			\node [etag, below=of sent1-1-4, yshift=-15mm, minimum height=7mm] (etag4)  [] {\color{black!80}\textsc{$E_{\scriptscriptstyle E,+}$}};
			\node [etag, below=of sent1-1-9, yshift=-15mm, minimum height=7mm] (etag9)  [] {\color{black!80}\textsc{$E_{\scriptscriptstyle S,0}$}};

			
			\foreach \from/\to in {{btag1/etag1},{etag1/atag1},{btag3/etag3},{etag4/atag4},{btag9/etag9},{etag9/atag9}}
			\draw[->,blue]   (\from) to[out=225,in=135] (\to);
			
			\foreach \from/\to in {{atag1/atag2},{etag3/etag4},{atag4/atag5},{atag5/atag6},{btag7/btag8},{btag8/btag9},{atag9/atag10}}
			\draw[->,blue]   (\from) to[out=0,in=180] (\to);
			
			\foreach \from/\to in {{atag2/btag3},{atag6/btag7}}
			\draw[->,red,line width=1.3pt]   (\from) to[out=0,in=200] (\to);

			\begin{pgfonlayer}{background}
			\node [scopebox, above=of sent1-1-1, xshift=4mm, yshift=-43mm, text height=5mm, minimum width=19mm] (s1)  [] {\color{blue!80}\textbf{\textsc{}}};
			\node [targetbox, above=of sent1-1-1, xshift=0mm, yshift=-7mm, text height=-2mm, minimum width=8mm] (e1)  [] {\color{blue!80}\textbf{\textsc{+}}};

			\node [scopebox, above=of sent1-1-4, xshift=9.5mm, yshift=-43mm, text height=-5mm, minimum width=54mm] (s2)  [] {\color{blue!80}\textbf{\textsc{}}};
			\node [targetbox, above=of sent1-1-3, xshift=5.5mm, yshift=-7mm, text height=-2mm, minimum width=22mm] (e2)  [] {\color{blue!80}\textbf{\textsc{+}}};

			\node [scopebox, above=of sent1-1-9, xshift=-5mm, yshift=-43mm, text height=-5mm, minimum width=42mm] (s3)  [] {\color{blue!80}\textbf{\textsc{}}};
			\node [targetbox, above=of sent1-1-9, xshift=0mm, yshift=-7mm, text height=-2mm, minimum width=11mm] (e3)  [] {\color{blue!80}\textbf{\textsc{0}}};
			
			\end{pgfonlayer}

			\end{tikzpicture} 
		}
		\caption{An alternative structured output for the same example with different sentiment boundaries.}
		\label{fig:ss_seq2}
	\end{figure}

The first sentiment span with positive ($+$) polarity contains only one word which is also the target. 
Such a single word target is also the beginning and the end of the target.
We use three tags $\mathbf{B}_+$, $\mathbf{E}_{S,+}$ and $\mathbf{A}_+$ to encode such information above.

The second sentiment span with positive ($+$) polarity contains a two-word target ``\textit{Shin Lim}''. 
The word ``and'' appearing before such target takes a tag $\mathbf{B}_+$. 
The words ``perform amazing magic'' appearing after such target take a tag $\mathbf{A}_+$ at each position.
As for the target, the word ``Shin'' at the beginning of the target takes tags $\mathbf{B}_+$ and $\mathbf{E}_{B,+}$, while the word ``Lim'' at the end of the target takes tags $\mathbf{E}_{E,+}$ and $\mathbf{A}_+$.

The third sentiment span with neutral ($0$) polarity contains a single-word target ``AGT''. 
Similarly, we use three tags $\mathbf{B}_0$, $\mathbf{E}_{S,0}$ and $\mathbf{A}_0$ to represent such single word target.
The word ``on'' appearing before such target takes a tag $\mathbf{B}_0$.
The word ``2018'' appearing afterwards takes a tag $\mathbf{A}_0$.

{\color{black}
Note that if there exists a target with length larger than 2, the tag $\mathbf{E}_{M,p}$ will be used. 
For example in Figure~\ref{fig:ss_seq1}, if the target phrase ``Shin Lim'' is replaced by ``Shin Bob Lim'', we will keep the tags at ``Shin'' and ``Lim'' unchanged. 
We assign a tag $\mathbf{E}_{M,+}$ at the word ``Bob'' to indicate that ``Bob'' appears in the middle of the target by following the \textit{BMES} tagging scheme.
}

Finally, we represent the label sequence by connecting adjacent tags sequentially with edges. 
Notice that for a given input sentence and the output targets as well as the associated targeted sentiment, there exist exponentially many possible label sequences, each specifying a different possible combinations of sentiment spans.
Figure~\ref{fig:ss_seq2} shows a label sequence for an alternative combination of the sentiment spans. 
Those label sequences representing the same input and output construct a latent variable in our model, capturing the flexible explicit structures in the output space.

We use a log-linear formulation to parameterize our model. 
Specifically, the probability of predicting a possible output $\vec{y}$, which is a list of targets and their associated sentiment information, given an input sentence $\vec{x}$, is defined as:
\begin{eqnarray}
p(\vec{y}|\vec{x}) = \frac{\sum_{\vec{h}}\exp{(s(\vec{x},\vec{y},\vec{h}))} }{\sum_{\vec{y}',\vec{h}'}\exp(s(\vec{x},\vec{y}^{'},\vec{h}^{'}) )}
\end{eqnarray}
where $s(\vec{x},\vec{y},\vec{h})$ is a score function defined over the sentence $\vec{x}$ and the output structure $\vec{y}$, together with the latent variable $\vec{h}$ that provides all the possible combinations of sentiment spans for the $(\vec{x,y})$ tuple. 
We define $E(\vec{x},\vec{y},\vec{h})$ as a set of all the edges appearing in all the label sequences for such combinations of sentiment spans.
To compute $s(\vec{x},\vec{y},\vec{h})$, we sum up the scores of each edge in $E(\vec{x},\vec{y},\vec{h})$:
$$s(\vec{x},\vec{y},\vec{h}) = \sum_{e \in E(\vec{x},\vec{y},\vec{h})} \phi_{\vec{x}}(e)$$
where $\phi_{\vec{x}}(e)$ is a score function defined over an edge $e$ for the input $\vec{x}$.

{\color{black}
The overall model is analogous to that of a neural CRF~\cite{NIPS2009_3869,do2010neural}; hence the inference and decoding follow standard marginal and MAP inference procedures.
For example, the prediction of $\vec{y}$ follows the Viterbi-like MAP inference procedure.
}

\subsection{Implicit Structure}
\begin{figure}[t!]
		\centering
		\adjustbox{max width=1.0\linewidth}{
			\begin{tikzpicture}[node distance=1.0mm and 1.0mm, >=Stealth, 
			wordnode/.style={draw=none, minimum height=5mm, inner sep=0pt},
			embnode/.style={draw=black, rounded corners, fill=red!20, minimum height=7mm, minimum width=30mm, inner sep=0pt},
			layernode/.style={draw=black, rounded corners, fill=red!20, minimum height=7mm, minimum width=55mm, inner sep=0pt},
			scoresnode/.style={draw=black, rounded corners, fill=red!40, minimum height=7mm, minimum width=18mm, inner sep=0pt},
			scorenode/.style={draw=black, fill=red!20, minimum height=5mm, minimum width=5mm, inner sep=0pt},
			chainLine/.style={line width=1pt,-, color=fontgray},
			scopebox/.style={draw=black, rounded corners, fill=yellow!20, minimum height = 24mm},
			scorebox/.style={draw=black,  rounded corners, fill=yellow!20, minimum height = 10mm, minimum width = 10mm},
			btag/.style={shape=circle, draw=black, rounded corners, fill=orange!20, minimum height=8mm, inner sep=0pt},
			atag/.style={shape=circle, draw=black, rounded corners, fill=orange!20, minimum height=8mm, inner sep=0pt},
			etag/.style={shape=circle, draw=black, rounded corners, fill=orange!20, minimum height=8mm, inner sep=0pt}
			]

			\node [embnode] (emb)  [] {\color{black!80}\textsc{$\vec{e}_k=[\vec{w}_k;\vec{c}_k]$}};
			
			\node [layernode, above=of emb, yshift=15mm, xshift=-30mm] (lstm)  [] {\color{black!80}\textsc{$\vec{h}_k=BiLSTM(\vec{e_1},\vec{e_2}, \cdots, \vec{e_n})$}};
			\node [layernode, above=of emb, yshift=15mm, xshift= 30mm] (att)  [] {\color{black!80}\textsc{$\vec{a}_k=SelfATT(\vec{e_1},\vec{e_2}, \cdots, \vec{e_n})$}};

			\node [scoresnode, above=of lstm, yshift=10mm, xshift=-15mm] (explicit_f_t)  [] {\color{black!80}\textsc{$f_t(\vec{h}_k)$}};
			\node [scoresnode, above=of lstm, yshift=10mm, xshift=15mm] (explicit_f_s)  [] {\color{black!80}\textsc{$f_s(\vec{h}_k)$}};
			\node [scoresnode, above=of att, yshift=10mm, xshift=0mm] (implicit_f_s)  [] {\color{black!80}\textsc{$g_s(\vec{a}_k)$}};

			\node [scorenode, above=of explicit_f_s, yshift=5mm, fill=red!20, xshift=0mm] (explicit_f_t_plus)  [] {\color{black!80}\textsc{$0$}};
			\node [scorenode, above=of explicit_f_s, yshift=5mm, fill=red!20, xshift=-5mm] (explicit_f_t_0)  [] {\color{black!80}\textsc{$+$}};
			\node [scorenode, above=of explicit_f_s, yshift=5mm, fill=red!20, xshift=5mm] (explicit_f_t_minus)  [] {\color{black!80}\textsc{$-$}};

			\node [scorenode, above=of explicit_f_t, yshift=5mm, fill=red!20, xshift=-7.5mm] (explicit_f_s_B)  [] {\color{black!80}\textsc{$B$}};
			\node [scorenode, above=of explicit_f_t, yshift=5mm, fill=red!20, xshift=-2.5mm] (explicit_f_s_M)  [] {\color{black!80}\textsc{$M$}};
			\node [scorenode, above=of explicit_f_t, yshift=5mm, fill=red!20, xshift=2.5mm] (explicit_f_s_E)  [] {\color{black!80}\textsc{$E$}};
			\node [scorenode, above=of explicit_f_t, yshift=5mm, fill=red!20, xshift=7.5mm] (explicit_f_s_S)  [] {\color{black!80}\textsc{$S$}};

			\node [scorenode, above=of implicit_f_s, yshift=5mm, fill=red!20, xshift=0mm] (implicit_f_s_plus)  [] {\color{black!80}\textsc{$0$}};
			\node [scorenode, above=of implicit_f_s, yshift=5mm, fill=red!20, xshift=-5mm] (implicit_f_s_0)  [] {\color{black!80}\textsc{$+$}};
			\node [scorenode, above=of implicit_f_s, yshift=5mm, fill=red!20, xshift=5mm] (implicit_f_s_minus)  [] {\color{black!80}\textsc{$-$}};


			\begin{pgfonlayer}{background}
			\node [scopebox, above=of explicit_f_t, xshift=0mm, yshift=-10mm, text height=5mm, minimum width=25mm] (s1)  [] {\color{blue!80}\textbf{\textsc{}}};
			\node [scopebox, above=of explicit_f_s, xshift=0mm, yshift=-10mm, text height=5mm, minimum width=25mm] (s2)  [] {\color{blue!80}\textbf{\textsc{}}};
			\node [scopebox, above=of implicit_f_s, xshift=0mm, yshift=-10mm, text height=5mm, minimum width=25mm] (s3)  [] {\color{blue!80}\textbf{\textsc{}}};

						
			\end{pgfonlayer}

			\draw[chainLine,->]   (emb) to[out=90,in=-90] (lstm);
			\draw[chainLine,->]   (emb) to[out=90,in=-90] (att);

			\draw[chainLine,->]   (lstm) to[out=90,in=-90] (s1);
			\draw[chainLine,->]   (lstm) to[out=90,in=-90] (s2);
			\draw[chainLine,->]   (att) to[out=90,in=-90] (s3);

			\end{tikzpicture} 
		}
		\caption{Neural Architecture}
		\label{fig:neural}
	\end{figure}
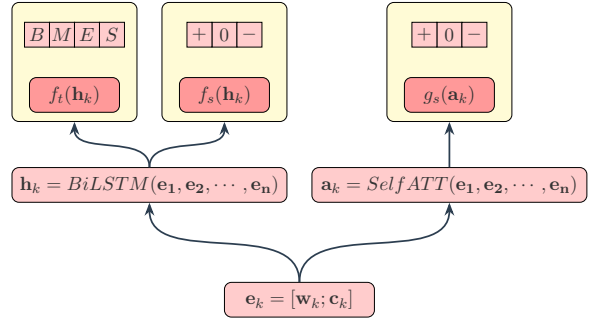

We propose a design for \textbf{EI} to efficiently learn rich implicit structures for exponentially many combinations of targets to predict.
To do so, we explain the process to assign scores to each edge $e$ from our neural architecture.
The three yellow boxes in Figure~\ref{fig:neural} compute scores for rich implicit structures from the neural architecture consisting of LSTM and self-attention.

Given an input token sequence $\vec{x}=\{x_1,x_2,\cdots,x_{n}\}$ of length $n$, we first compute the concatenated embedding $\vec{e}_k=[\vec{w}_k;\vec{c}_k]$ based on word embedding $\vec{w}_k$ and character embedding $\vec{c}_k$ at position $k$.


As illustrated on the left part in Figure~\ref{fig:neural}, we then use a Bi-directional LSTM to encode context features and obtain hidden states $\vec{h}_k=\mathrm{BiLSTM}(\vec{e_1},\vec{e_2}, \cdots, \vec{e_n})$. 
We use two different linear layers $f_t$ and $f_s$ to compute scores for target and sentiment respectively. 
The linear layer $f_t$ returns a vector of length $4$, with each value in the vector indicating the score of the corresponding tag under the \textit{BMES} tagging scheme. 
The linear layer $f_s$ returns a vector of length $3$, with each value representing the score of a certain polarity of $+,0,-$. 
We assign such scores to each type of edge as follows:
$$\phi_{\vec{x}}(\mathbf{E}^{k}_{\epsilon,p}\mathbf{E}^{k+1}_{\epsilon',p})=f_{t}(\vec{h}_k)_{\epsilon}$$
$$\phi_{\vec{x}}(\mathbf{E}^{k}_{\epsilon,p}\mathbf{A}^{k}_{p})=f_{t}(\vec{h}_k)_{\epsilon}$$
$$\phi_{\vec{x}}(\mathbf{B}^{k}_{p}\mathbf{B}^{k+1}_{p})=f_{s}(\vec{h}_k)_{p}$$
$$\phi_{\vec{x}}(\mathbf{A}^{k}_{p}\mathbf{A}^{k+1}_{p})=f_{s}(\vec{h}_k)_{p}$$
$$\phi_{\vec{x}}(\mathbf{A}^{k}_{p}\mathbf{B}^{k+1}_{p'})=f_{s}(\vec{h}_k)_{p}$$
Note that the subscript $p$ and $\epsilon$ at the right hand side of above equations denote the corresponding index of the vector that $f_t$ or $f_s$ returns. 
We apply $f_{t}$ on edges $\mathbf{E}^{k}_{\epsilon,p}\mathbf{E}^{k+1}_{\epsilon',p}$ and $\mathbf{E}^{k}_{\epsilon,p}\mathbf{A}^{k}_{p}$, since words at these edges are parts of the target phrase in a sentiment span.
Similarly, we apply $f_{s}$ on edges $\mathbf{B}^{k}_{p}\mathbf{B}^{k+1}_{p}$,$\mathbf{A}^{k}_{p}\mathbf{A}^{k+1}_{p}$ and $\mathbf{A}^{k}_{p}\mathbf{B}^{k+1}_{p'}$, since words at these edges contribute the sentiment information for the target in the sentiment span.



%

As illustrated in Figure~\ref{fig:neural}, we calculate $\vec{a}_k$, the output of self-attention at position $k$: 
$$\vec{a}_k=\sum_{j=1}^{n} \alpha_{k,j} \vec{e}_j$$
$$\alpha_{k,j}= \softmax_{j}(\vec{\beta}_{k,j})$$
$$\vec{\beta}_{k,j} = U^{T} \mathrm{ReLu}(W [\vec{e}_k; \vec{e}_j] + b)$$
where $\alpha_{k,j}$ is the normalized weight score for $\vec{\beta}_{k,j}$, and $\vec{\beta}_{k,j}$ is the weight score calculated by target representation at position $k$ and contextual representation at position $j$.
In addition, $W$ and $b$ as well as the attention matrix $U$ are the weights to be learned. 
Such a vector $\vec{a}_k$ encodes the implicit structures between the word $x_k$ and each word in the remaining sentence.

Motivated by the character embeddings~\cite{lample2016neural} which are generated based on hidden states at two ends of a subsequence, we encode such implicit structures for a target similarly.
For any target starting at the position $k_1$ and ending at the position $k_2$, we could use $\vec{a}_{k_1}$ and $\vec{a}_{k_2}$ at two ends to represent the implicit structures of such a target.
We encode such information on the edges $\mathbf{B}^{k_1}_{p}\mathbf{E}^{k_1}_{\epsilon,p}$ and $\mathbf{E}^{k_2}_{\epsilon,p}\mathbf{A}^{k_2}_{p}$ which appear at the beginning and the end of a target phrase respectively with sentiment polarity $p$. 
To do so, we assign the scores calculated from the self-attention to such two edges: 
$$\phi_{\vec{x}}(\mathbf{B}^{k_1}_{p}\mathbf{E}^{k_1}_{\epsilon,p})=g_{s}(\vec{a}_{k_1})_{p}$$
$$\phi_{\vec{x}}(\mathbf{E}^{k_2}_{\epsilon,p}\mathbf{A}^{k_2}_{p}) \pluseq g_{s}(\vec{a}_{k_2})_{p}$$
where $g_{s}$ returns a vector of length $3$ with scores of three polarities. 

Note that $\vec{h}_k$ and $\vec{a}_k$ could be pre-computed at every position $k$ and assigned to the corresponding edges.
Such an approach allows us to maintain the inference time complexity $O(Tn)$, where $T$ is the maximum number of tags at each position which is $9$ in this work and $n$ is the number of words in the input sentence.
This approach enables \textbf{EI} to efficiently learn rich implicit structures from LSTM and self-attention for exponentially many combinations of targets.

\section{Experimental Setup}
\subsection*{Data}
We mainly conduct our experiments on the datasets released by~\citet{mitchell2013open}. 
They contain 2,350 English tweets and 7,105 Spanish tweets, with target and targeted sentiment annotated. 
See Table \ref{fig:training_data_stats} for corpus statistics.

\begin{table}[t]
\centering

\begin{subtable}{.5\textwidth}
\centering
\begin{small}
\begin{tabular}{l  c c c c}
\toprule
 & \#Target &  \#$+$ & \#$-$ & \#$0$ \\
\midrule
English & 3,288 &  707 & 275 & 2,306 \\
Spanish & 6,658 &  1,555 & 1,007 & 4,096 \\
\bottomrule
\end{tabular}
\end{small}
\vspace{-2mm}
\caption{Statistics on polarity of named entities}
\vspace{1mm}
\end{subtable}


\begin{subtable}{.5\textwidth}
\centering
\begin{small}
\begin{tabular}{l  c c c c}
\toprule
Target length & 1 & 2 & 3 & $>=4$ \\
\midrule
English & 1,910 &  1,032 & 232 &  114 \\
Spanish & 4,201 &  1,794 & 417 & 246 \\
\bottomrule
\end{tabular}
\end{small}
\vspace{-2mm}
\caption{Statistics on target length}
\vspace{1mm}
\end{subtable}

\begin{subtable}{.5\textwidth}
\centering
\begin{small}
\begin{tabular}{l  c c c c c}
\toprule
\#Target & 1 & 2 & 3 &  $>=4$ \\
\midrule
English & 1,692 &  465 & 135 &  58\\
Spanish & 3,855 &  903 & 221 & 69 \\
\bottomrule
\end{tabular}
\end{small}
\vspace{-2mm}
\caption{Statistics on number of targets per sentence}
\vspace{0mm}
\end{subtable}
\caption{Corpus Statistics of Main Dataset}
\label{fig:training_data_stats}
\end{table}

\subsection*{Evaluation Metrics}
Following the previous works, we report the \textit{precision} ($P.$), \textit{recall} ($R.$) and \textit{$F_1$} scores for target recognition and targeted sentiment. 
Note that a correct target prediction requires the boundary of the target to be correct, and a correct targeted sentiment prediction requires both target boundary and sentiment polarity to be correct. 

\begin{table*}[t!]
	\centering
	 \begin{adjustwidth}{0.0cm}{0.0cm}
	\scalebox{0.6}{
		\begin{tabular}{l|lc|cccccc|cccccc}
			\toprule
			\multirow{2}{*}{Model}& \multicolumn{2}{c}{Structure} & \multicolumn{6}{c}{English} & \multicolumn{6}{c}{Spanish}  \\ 
			&  &  & \multicolumn{3}{c}{Target Recognition} & \multicolumn{3}{c|}{Targeted Sentiment} & \multicolumn{3}{c}{Target Recognition} & \multicolumn{3}{c}{Targeted Sentiment}  \\ 
			
			 & \textit{Explicit} & \textit{Implicit} & $P.$ & $R.$ & $F_1$& $P.$ & $R$. & $F_1$& $P.$ & $R.$ & $F_1$& $P.$ & $R.$ & $F_1$\\
			 \midrule

			 Pipeline~\cite{zhang2015neural} & \textit{{\color{white}fix}fixed} & \textit{MLP} + \textit{discrete + emb} & 60.69& 51.63& 55.67& 43.71 & 37.12 & 40.06 & 70.23 & 62.00 & 65.76 & 45.99 & 40.57 & 43.04 \\
			  Joint~\cite{zhang2015neural}& \textit{{\color{white}fix}fixed} &  \textit{MLP} + \textit{discrete + emb} & 61.47 & 49.28 & 54.59 & 44.62 & 35.84 & 39.67 & 71.32& 61.11 & 65.74 & 46.67 & 39.99 & 43.02\\
			 Collapse~\cite{zhang2015neural}&\textit{{\color{white}fix}fixed} & \textit{MLP} + \textit{discrete + emb} & 63.55 & 44.98 & 52.58 & 46.32 & 32.84 & 38.36 & 73.51 & 53.30 & 61.71 & 47.69 & 34.53 & 40.00\\
			 Bi-GRU~\cite{ma2018joint} & \textit{{\color{white}fix}fixed} & \textit{GRU} + \textit{emb} &  58.13 & 43.46 & 49.62 & 45.76 & 32.29 & 37.73 & 65.24 & 53.02 & 58.45 & 46.33 & 37.50 & 41.45\\
			 MBi-GRU~\cite{ma2018joint} & \textit{{\color{white}fix}fixed} & \textit{MGRU} + \textit{emb} &  58.27 & 49.01 & 53.24 &  45.80 &  35.21 & 39.81 & 66.14 & 60.07 &  62.95 & 45.61 & 40.04 & 42.64\\
			 HBi-GRU~\cite{ma2018joint} & \textit{{\color{white}fix}fixed} & \textit{GRU} + \textit{emb + char} &  57.24 & 53.88 & 55.41 & 44.94 & 38.60 & 41.52 & 68.24 & 61.81 & 64.82 & 46.53 & 42.21 & 44.18\\
			 HMBi-GRU~\cite{ma2018joint} &\textit{{\color{white}fix}fixed} & \textit{MGRU} + \textit{emb + char} &  60.12 & 53.68 & 56.98 & 46.52 & 39.99 & 42.87 & 68.64 & 63.66 & 66.01 & 48.09 & 43.44 & 45.61\\
			 
			 SS~\cite{li2017sentimentscope}& \textit{flexible} & \textit{discrete}  & 63.18 & 51.67 & 56.83 & 44.57 & 36.48 & 40.11 & 71.49 & 61.92 & 66.36 & 46.06 & 39.89 & 42.75 \\
			 SS + \textit{emb}~\cite{li2017sentimentscope}& \textit{flexible} & \textit{discrete + emb} & 66.35 & 56.59 & 61.08 & 47.30 & 40.36 & 43.55 & 73.13 & 64.34 & 68.45 & 47.14 & 41.48 & 44.13 \\
			 \hdashline
			  SA-CRF & \textit{{\color{white}fix}fixed} & \textit{LSTM} + \textit{SA} + \textit{emb + char}  & 60.26 & 55.60 & 57.53 & 42.95 & 40.46 & 41.45 & 68.47 & 66.39 & 67.26 & 42.22 & 42.97 & 42.47 \\
			  {\color{black} \textbf{E-I}} &  \textit{flexible} & \textit{LSTM} + \textit{SA} +  \textit{emb + char} & 67.11 & 58.37 & 62.34 & 47.47 & 41.31 & 44.11 & 73.47 & 65.91 & 69.44 & 47.80 & 42.90 & 45.19 \\
			 \textbf{EI-} & \textit{flexible} & \textit{LSTM} + \textit{emb + char} & 68.67 & 57.52 & 62.54 &  48.73 & 40.89 & 44.42 & 72.62 & 66.97 & 69.61 & 47.06 & 43.45 & 45.14 \\
			 \textbf{EI} & \textit{flexible} & \textit{LSTM} + \textit{SA} +  \textit{emb + char} & {\bf 69.70} & {\bf 58.33} & {\bf 63.48} & {\bf 49.78} & {\bf 41.71} & {\bf 45.37} & {\bf 74.25} & {\bf 68.37} & {\bf 71.17} & {\bf 48.10} & {\bf 44.29}& {\bf 46.11} \\

			\bottomrule
		\end{tabular}
	}
	\end{adjustwidth}
	\caption{Main Results. \textit{fixed} stands for chain structures and \textit{flexible} for latent structures. \textit{discrete}, \textit{emb} and   \textit{char} denote discrete features, word embeddings and character embeddings respectively. \textit{SA} represents self-attention.}
	\label{tab:mainresult}
\end{table*}

\subsection*{Hyperparameters}
We adopt pretrained embeddings from ~\citet{pennington2014glove} and \citet{cieliebak-etal-2017-twitter} for English data and Spanish data respectively.
We use a 2-layer LSTM (for both directions) with a hidden dimension of 500 and 600\footnote{{\color{black}We use a larger LSTM hidden size for Spanish since dimension of Spanish word embedding (200) is larger than dimension of English word embedding (100).}} for English data and Spanish data respectively. 
The dimension of the attention weight $U$ is 300. 
As for optimization, we use the Adam~\cite{kingma2014adam} optimizer to optimize the model with batch size 1 and dropout rate $0.5$. 
All the {\color{black}neural weights} are initialized by Xavier~\cite{glorot2010understanding}.

\subsection*{Training and Implementation}
We train our model for a maximal of 6 epochs.
We select the best model parameters based on the best $F_1$ score on the development data after each epoch.
Note that we split $10\%$ of data from the training data as the development data\footnote{Detailed split information is released with our code.}.
The selected model is then applied to the test data for evaluation. 
During testing, we map words not appearing in the training data to the \textit{UNK} token.
Following the previous works, we perform 10-fold cross validation and report the average results.
Our models and variants are implemented using PyTorch~\cite{paszke2017automatic}. 

\subsection*{Baselines}
We consider the following baselines:
\begin{itemize}
\item {Pipeline~\cite{zhang2015neural} and Collapse~\cite{zhang2015neural}} both are linear-chain CRF models using discrete features and embeddings. The former predicts targets first and calculate targeted sentiment for each predicted target. The latter outputs a tag at each position by collapsing the target tag and sentiment tag together.
\item {Joint~\cite{zhang2015neural}} is a linear-chain SSVM model using both discrete features and embeddings. Such a model jointly produces target tags and sentiment tags.
\item {Bi-GRU~\cite{ma2018joint} and MBi-GRU~\cite{ma2018joint} } are both linear-chain CRF models using word embeddings. The former uses bi-directional GRU and the latter uses multi-layer bi-directional GRU.
\item {HBi-GRU~\cite{ma2018joint} and HMBi-GRU~\cite{ma2018joint}} are both linear-chain CRF models using word embeddings and character embedding. The former uses bi-directional GRU and the latter uses multi-layer bi-directional GRU.
\item {SS~\cite{li2017sentimentscope} and SS + \textit{emb}~\cite{li2017sentimentscope}} are both based on a latent CRF model to learn flexible explicit structures. The former uses discrete features and the latter uses both discrete features and word embeddings.
\item {SA-CRF} is a linear-chain CRF model with self-attention. Such a model concatenates the hidden state from LSTM and a vector constructed by self-attention at each position, and feeds them into CRF as features. The model attempts to capture rich implicit structures in the input space, {\color{black}but it does not put effort on explicit structures in the output space.}
\item {\color{black} {\textbf{E-I}}  is a weaker version of \textbf{EI}. Such a model removes the \textit{BMES} sub-tags in the \textbf{E} tag, causing the model to learn less explicit structural information in the output space. }
\item {\textbf{EI-}}  is a weaker version of \textbf{EI}. Such a model removes the self-attention from \textbf{EI}, causing the model to learn less expressive implicit structures in the input space.




\end{itemize}

\section{Results and Discussion}
\subsection{Main Results}
The main results are presented in Table~\ref{tab:mainresult}, where explicit structures as well as implicit structures are indicated for each model for clear comparisons. 


In general, our model \textbf{EI} outperforms all the baselines. {\color{black}Specifically}, it outperforms the strongest baseline \textbf{EI-} significantly with $p < 0.01$ on the English and Spanish datasets in terms of $F_1$ scores\footnote{We have conducted significance test using the bootstrap resampling method~\cite{koehn-2004-statistical}.}.
Note that \textbf{EI-} which models flexible explicit structures and less implicit structural information, achieves better performance than most of the baselines, indicating flexible explicit structures contribute a lot to the performance boost.


Now let us take a closer look at the differences based on detailed comparisons. First of all, we compare our model \textbf{EI} with the work proposed by~\citet{zhang2015neural}. 
The Pipeline model (based on CRF) as well as Joint and Collapse models (based on SSVM) in their work capture \textit{fixed} explicit structures.
Such two models rely on multi-layer perceptron (MLP) to obtain the local context features for implicit structures.
These two models do not put much effort to capture better explicit structures and implicit structures. 
Our model \textbf{EI} {\color{black}(and even \textbf{EI-}) outperforms} these two models significantly.
We also compare our work with models in~\citet{ma2018joint}, which also capture \textit{fixed} explicit structures.
Such models leverage different GRUs (single-layer or multi-layer) and different input features (word embeddings and character representations) to learn better contextual features.
Their best result by HMBi-GRU is obtained with multi-layer GRU with word embeddings and character embeddings.
As we can see, our model \textbf{EI} outperforms HMBi-GRU {\color{black}under all evaluation metrics}.
On the English data, \textbf{EI} obtains $6.50$ higher $F_1$ score and $2.50$ higher $F_1$ score on target recognition and targeted sentiment respectively.
On Spanish, \textbf{EI} obtains $5.16$ higher $F_1$ score and $0.50$ higher $F_1$ score  on target recognition and targeted sentiment respectively. 
Notably, compared with HMBi-GRU, even \textbf{EI-} capturing the flexible explicit structures achieves better performance on most of metrics and obtains the comparable results in terms of precision and $F_1$ score on Spanish.
Since both \textbf{EI} and \textbf{EI-} models attempt to capture the \textit{flexible} explicit structures, the comparisons above imply the importance of modeling such \textit{flexible} explicit structures in the output space. 

We also compare \textbf{EI} with \textbf{E-I}. 
The difference between these two models is that \textbf{E-I} removes the \textit{BMES} sub-tags. 
Such a model captures less explicit structural information in the output space. 
We can see that \textbf{EI} outperforms \textbf{E-I}.
Such results show that adopting \textit{BMES} sub-tags in the output space to capture explicit structural information is beneficial.

Now we compare \textbf{EI} with SA-CRF which is a linear-chain CRF model with self-attention.
Such a model attempts to capture rich implicit structures, and \textit{fixed} explicit structures.
The difference between \textbf{EI} and SA-CRF is that our model \textbf{EI} captures \textit{flexible} explicit structures in the output space which model output representations as latent variables.
We can see that \textbf{EI} outperforms SA-CRF on all the metrics. 
Such a comparison also implies the importance of capturing \textit{flexible} explicit structures in the output space.

Next, we focus on the comparisons with SS~\cite{li2017sentimentscope} and SS + \textit{emb}~\cite{li2017sentimentscope}. 
Such two models as well as our models all capture the \textit{flexible} explicit structures.
As for the difference, both two SS models rely on hand-crafted discrete features to capture implicit structures, while our model \textbf{EI} and \textbf{EI-} learn better implicit structures by LSTM and self-attention.
Furthermore, our models only require word embeddings and character embeddings as the input to our neural architecture to model rich implicit structures, leading to a comparatively simpler and more straightforward design. 
The comparison here suggests that LSTM and self-attention neural networks are able to capture better implicit structures than hand-crafted features.

\begin{table}[t!]
\centering
\scalebox{0.7}{
\begin{tabular}{l|ccc|ccc}
	\toprule
   \multirow{2}{*}{Model}  	 & \multicolumn{3}{c|}{Subj (+/-,o)} & \multicolumn{3}{c}{{\color{black}SA} (+,-)} \\
    &  $P.$  & $R.$  & $F_1$ & $P.$  & $R.$  & $F_1$  \\
    \midrule
    {\small \citet{zhang2015neural}} & {49.2} & 42.1 & 45.3 & {40.9} & 21.6 & 27.9 \\
	{\small SS + \textit{emb}~\cite{li2017sentimentscope}}& {50.0} & {44.0} & {46.8} & {37.6} & {25.4} & {30.2} \\
	{\small SA-CRF} & 44.8 & 45.2 & 44.9 & 35.2 & 25.6 & 29.3 \\
	{\small \textbf{EI-}} & 49.7 & 45.8 & 47.6 & {\bf 43.0} & 24.9 & 30.2 \\
	{\small \textbf{EI}} & {\bf 50.5} & {\bf 46.5} & {\bf 48.4} & 42.0 & {\bf 25.6} & {\bf 31.5} \\
	\bottomrule
\end{tabular}
}
\caption{\color{black} Results on subjectivity as well as non-neutral sentiment analysis on the Spanish dataset. Subj(+/-,o): subjectivity for all polarities. SA(+,-): sentiment analysis for non-neutral polarities. }
\label{tab:subjectivity}
\end{table}

Finally, we compare \textbf{EI} with \textbf{EI-}. 
We can see that the $F_1$ scores of targeted sentiment for both English and Spanish produced by \textbf{EI} are $0.95$ and $0.97$ points higher than \textbf{EI-}. 
The main difference here is that \textbf{EI} makes use of self-attention to capture richer implicit structures between each target phrase and all words in the complete sentence. 
The comparisons here indicate the importance of capturing rich implicit structures using self-attention on this task.

\subsubsection*{Robustness}

Overall, all these comparisons above based on empirical results show the importance of capturing both \textit{flexible} explicit structures in the output space and rich implicit structures by LSTM and self-attention in the input space. 

\begin{figure}[t!]
 \centering
 \begin{adjustwidth}{0.5cm}{0.0cm}
 \scalebox{1.0}
  {
  \includegraphics[width=\linewidth]{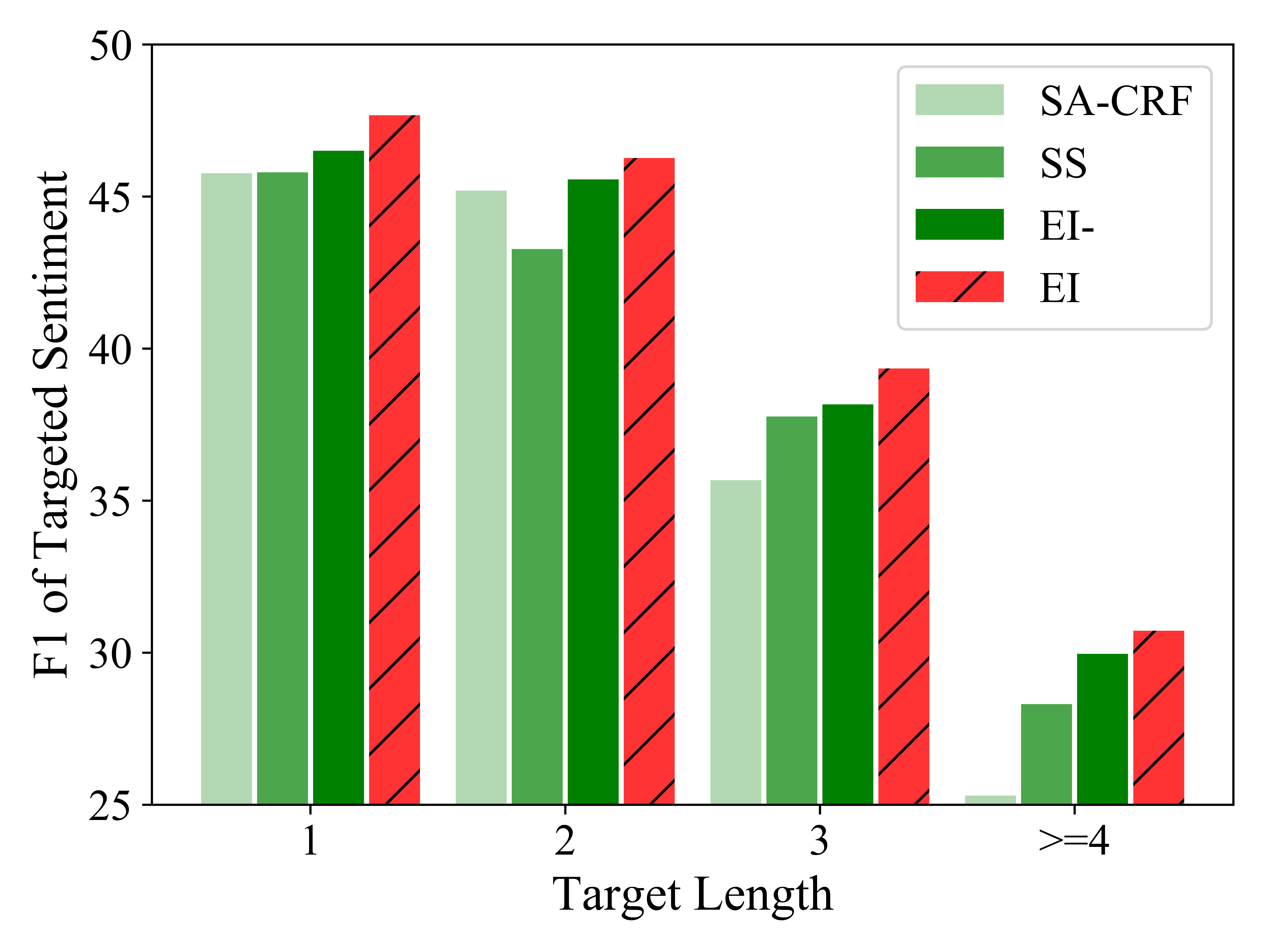}
  }
    \caption{Results of different lengths on Spanish}
  \label{fig:chunklength}
  \end{adjustwidth}
\end{figure}

We analyze the model robustness by assessing the performance on the targeted sentiment for targets of different lengths. 
For both English and Spanish, we group targets into 4 categories respectively, namely length of $1$, $2$, $3$ and $\geq 4$. 
Figure~\ref{fig:chunklength} reports the $F_1$ scores of targeted sentiment for such 4 groups on Spanish\footnote{See the English results in Figure~\ref{fig:chunklength_en} in the appendix.}. See the English results in the supplementary material.
As we can see \textbf{EI} outperforms all the baselines on all groups. 

Furthermore, following the comparisons in~\citet{zhang2015neural}, we also measure the precision, recall and $F_1$ of subjectivity and non-neutral polarities on the Spanish dataset.
Results are reported in Table~\ref{tab:subjectivity}\footnote{Only Spanish results are available in~\citet{zhang2015neural}.}. 
The subjectivity measures whether a target phrase expresses an opinion or not according to \citet{liu2010sentiment}. 
Comparing with the best-performing system's results reported in~\citet{zhang2015neural} and~\citet{li2017sentimentscope}, our model \textbf{EI} can achieve higher $F_1$ scores on subjectivity and non-neutral polarities.


\subsubsection*{Error Analysis}
We conducted error analysis for our main model \textbf{EI}. 
We calculate $F_1$ scores based on the partial match instead of exact match. 
The $F_1$ scores for target partial match is $76.04$ and $83.82$ for English and Spanish respectively. 
We compare these two numbers against $63.48$ and $71.17$ which are the $F_1$ scores based on exact match. 
This comparison indicates that boundaries of many {\color{black}predicted targets do not match exactly with those of the correct targets}.
Furthermore, we investigate the errors caused by incorrect sentiment polarities. 
We found that the major type of errors is to incorrectly predict positive targets as neutral targets.
Such errors contribute $64\%$ and $36\%$ of total errors for English and Spanish respectively. 
We believe they are mainly caused by challenging expressions in the tweet input text.
Such challenging expressions such as “\textit{below expectations}” are very sparse in the data, which makes effective learning for such phrases difficult.

\begin{table*}[t!]
	\centering
	\scalebox{0.75}{
		\begin{tabular}{l|cccccc|cccccc}
			\toprule
			\multirow{3}{*}{Model}& \multicolumn{6}{c}{English} & \multicolumn{6}{c}{Spanish}  \\ 
			& \multicolumn{3}{c}{Target Recognition} & \multicolumn{3}{c|}{Targeted Sentiment} & \multicolumn{3}{c}{Target Recognition} & \multicolumn{3}{c}{Targeted Sentiment} \\ 
			 & Prec. & Rec. & $F_1$& Prec. & Rec. & $F_1$& Prec. & Rec. & $F_1$& Prec. & Rec. & $F_1$\\
			 \midrule
		     \textbf{EI} & {\bf 69.70} & {\bf 58.33} & {\bf 63.48} & {\bf 49.78} & {\bf 41.71} & {\bf 45.37} & {\bf 74.25} & {\bf 68.37} & {\bf 71.17} & {\bf 48.10} & {\bf 44.29}& {\bf 46.11} \\
			\textbf{EI} (\textit{i:MLP})& 64.47 & 56.58 & 60.20 & 46.23 & 40.48 & 43.12 & 70.95 & 65.80 & 68.27 & 43.64 & 40.46 & 41.98\\
			\textbf{EI} (\textit{i:Identity}) & 63.24 & 55.73 & 59.20 & 45.10 & 39.79 & 42.24 & 69.38 & 66.27 & 67.77 & 43.66 & 41.68 & 42.63\\
			 \textbf{EI-}  & 68.67 & 57.52 & 62.54 &  48.73 & 40.89 & 44.42 & 72.62 & 66.97 & 69.61 & 47.06 & 43.45 & 45.14 \\
		
		
			\bottomrule
		\end{tabular}
	}
	\caption{Effect of Implicit Structures}
	\label{tab:model_variants}
\end{table*}

\subsection{Effect of Implicit Structures}
In order to understand whether the implicit structures are truly making contributions in terms of the overall performance, we compare the performance among four models: \textbf{EI} and \textbf{EI-} as well as two variants \textbf{EI} (\textit{i:MLP}) and \textbf{EI} (\textit{i:Identity}) (where \textit{i} indicates the implicit structure). 
Such two variants replace the implicit structure by other components: 
\begin{itemize}
 \item{\textbf{EI} (\textit{i:MLP})} replaces self-attention by multi-layer perceptron (MLP) for implicit structures. Such a variant attempts to capture implicit structures for a target phrase towards words restricted by a window of size $3$ centered at the two ends of the target phrase. 
 \item{\textbf{EI} (\textit{i:Identity})} replaces self-attention by an identity layer\footnote{The identity layer returns the identical input data.} as implicit structure. Such a variant attempts to capture implicit structures for a target phrase towards words at the two ends of the target phrase exactly.
\end{itemize}

\begin{figure}[t!]
		\centering
		\adjustbox{max width=1.0\linewidth}{
			\begin{tikzpicture}[node distance=1.0mm and 1.0mm, >=Stealth, 
			wordnode/.style={draw=none, minimum height=5mm, inner sep=0pt},
			chainLine/.style={line width=1pt,-, color=fontgray},
			scopebox/.style={draw=black, rounded corners, fill=yellow!20, dashed, minimum height = 15mm},
			targetbox/.style={draw=black, rounded corners, fill=red!20, minimum height = 10mm},
			btag/.style={shape=circle, draw=black, rounded corners, fill=orange!20, minimum height=8mm, inner sep=0pt},
			atag/.style={shape=circle, draw=black, rounded corners, fill=orange!20, minimum height=8mm, inner sep=0pt},
			etag/.style={shape=circle, draw=black, rounded corners, fill=orange!20, minimum height=8mm, inner sep=0pt}
			]

			\matrix (sent1) [matrix of nodes, nodes in empty cells, execute at empty cell=\node{\strut};]
			{
				\textbf{Czech} & [1mm] \textbf{Republic} &[1mm] , & [1mm] \textbf{Greece} &  [1mm]  and & [1mm] \textbf{Russian} & [1mm] ... & [1mm] sound & [1mm] good \\
			};

			\begin{pgfonlayer}{background}
			\node [scopebox, above=of sent1-1-1, xshift=9.5mm, yshift=-10mm, text height=5mm, minimum width=30mm] (s1)  [] {\color{blue!80}\textbf{\textsc{}}};
			\node [targetbox, above=of sent1-1-1, xshift=9.5mm, yshift=-7mm, text height=-2mm, minimum width=29mm] (e1)  [] {{\color{blue!80}\textbf{\textsc{+}}} / {\color{red!80}\textbf{\textsc{0}}}};

			\node [scopebox, above=of sent1-1-4, xshift=-2mm, yshift=-10mm, text height=-5mm, minimum width=19.5mm] (s2)  [] {\color{blue!80}\textbf{\textsc{}}};
			\node [targetbox, above=of sent1-1-4, xshift=0mm, yshift=-7mm, text height=-2mm, minimum width=14.5mm] (e2)  [] {{\color{blue!80}\textbf{\textsc{+}}} / {\color{red!80}\textbf{\textsc{+}}}};

			\node [scopebox, above=of sent1-1-6, xshift=11mm, yshift=-10mm, text height=-5mm, minimum width=55mm] (s3)  [] {\color{blue!80}\textbf{\textsc{}}};
			\node [targetbox, above=of sent1-1-6, xshift=0mm, yshift=-7mm, text height=-2mm, minimum width=16mm] (e3)  [] {{\color{blue!80}\textbf{\textsc{+}}} / {\color{red!80}\textbf{\textsc{+}}}};
			
			\end{pgfonlayer}

			\end{tikzpicture} 
		}
		\caption{An example sentence in the test data.}
		\label{fig:qualitative}
	\end{figure}
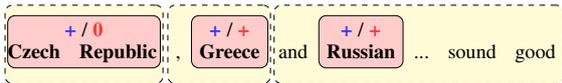

Overall, those variants perform worse than \textbf{EI} on all the metrics. 
When the self-attention is replaced by MLP or the identity layer for implicit structures, the performance drops a lot on both target and targeted sentiment. 
Such two variants \textbf{EI} (\textit{i:MLP}) and \textbf{EI} (\textit{i:Identity}) consider the words within a small window centered at the two ends of the target phrase, which might not be capable of capturing the desired implicit structures. 
The \textbf{EI-} model capturing less implicit structural information achieves worse results than \textbf{EI}, but obtains better results than the two variants discussed above. 
This comparison implies that properly capturing implicit structures as the complement of explicit structural information is essential.

\subsection{Qualitative Analysis}
We present an example sentence in the test data in Figure~\ref{fig:qualitative}, where {\color{black}the gold targets are in bold, the predicted targets are in the pink boxes, the gold sentiment is in blue and predicted sentiment is in red.
}
\textbf{EI} makes all correct predictions for three targets.
\textbf{EI-} predicts correct boundaries for three targets and the targeted sentiment predictions are highlighted in Figure~\ref{fig:qualitative}.
As we can see, \textbf{EI-} incorrectly predicts the targeted sentiment on the first target as neural ($0$).
The first target here is far from the sentiment expression ``\textit{sound good}'' which is not in the first sentiment span, making \textbf{EI-} not capable of capturing such a sentiment expression.
This qualitative analysis helps us to better understand the importance to capture implicit structures using both LSTM and self-attention.

\subsection{Additional Experiments}
We also conducted experiments on multi-lingual Restaurant datasets from SemEval 2016 Task 5~\cite{pontiki2016semeval}, where aspect target phrases and aspect sentiments are provided.  \footnote{See the data statistics in Table~\ref{fig:training_add_data_stats} in the appendix.} 
We regard each aspect target phrase as a target and assign such a target with the corresponding aspect sentiment polarity in the data.
Note that we remove all the instances which contain no targets in the training data.
Following the main experiment, we split $10\%$ of training data as development set for the selection of the best model during training.

We report the $F_1$ scores of target and targeted sentiment for English, Dutch and Russian\footnote{We use the pretrained embedding for Dutch and Russian from \url{https://github.com/Kyubyong/wordvectors}.} respectively in Table~\ref{tab:additional}. 
The results show that \textbf{EI} achieves the best performance.
The performance of SS ~\cite{li2017sentimentscope} is much worse on Russian due to the inability of discrete features in SS to capture the complex morphology in Russian.

\section{Related Work}


We briefly survey the research efforts on two types of TSA tasks mentioned in the introduction.
Note that TSA is related to aspect sentiment analysis which is to determine the sentiment polarity given a target and an aspect describing a property of related topics. 


\subsubsection*{\textit{Predicting sentiment for a given target}}
Such a task is typically solved by leveraging sentence structural information, such as syntactic trees~\cite{dong2014adaptive}, dependency trees~\cite{wang2016recursive} as well as surrounding context based on LSTM~\cite{tang2016target}, GRU~\cite{zhang2016gated} or CNN~\cite{xue2018aspect}. 
Another line of works leverage self-attention~\cite{liu2017EACL} or memory networks~\cite{tang2016aspect} to encode rich global context information.
\citet{wang2018learning} adopted the segmental attention~\cite{kong2015segmental} to model the important text segments to compute the targeted sentiment.
\citet{P18-1088} studied the issue that the different combinations of target and aspect may result in different sentiment polarity. 
They proposed a model to distinguish such different combinations based on memory networks to produce the representation for aspect sentiment classification.
\subsubsection*{\textit{Jointly predicting targets and their associated sentiment }}
{\color{black}Such a joint task is usually regarded as sequence labeling problem.}
\citet{mitchell2013open} introduced the task of open domain targeted sentiment analysis.
They proposed several models based on CRF such as the pipeline model, the collapsed model as well as the joint model to predict both targets and targeted sentiment information. 
{\color{black} Their experiments showed that the collapsed model and the joint model could achieve better results, implying the benefit of the joint learning on this task.}
\citet{zhang2015neural} proposed an approach based on structured SVM~\cite{taskar2005learning,tsochantaridis2005large} integrating both discrete features and neural features for this joint task.
\citet{li2017sentimentscope} proposed the sentiment scope model motivated from a linguistic phenomenon to represent the structure information for both the targets and their associated sentiment polarities. 
They modelled the latent sentiment scope based on CRF with latent variables, and achieved the best performance among all the existing works.
{\color{black} However, they did not explore much on the implicit structural information and their work mostly relied on hand-crafted discrete features.}
\citet{ma2018joint} adopted a multi-layer GRU to learn targets and sentiments jointly by producing the target tag and the sentiment tag at each position. They introduced a constraint forcing the sentiment tag at each position {\color{black}to be consistent} with the target tag. 
{\color{black} However, they did not explore the explicit structural information in the output space as we do in this work.}

\begin{table}[t!]
\centering
\scalebox{0.70}{
\begin{tabular}{l|cc|cc|cc}
	\toprule
   \multirow{2}{*}{\small Model}  	 & \multicolumn{2}{c|}{\small English} &   \multicolumn{2}{c|}{\small Dutch}   &   \multicolumn{2}{c}{\small Russian}  \\
    &  \textit{{\small target}}  & \textit{{\small sent}}  & \textit{{\small target}}  & \textit{{\small sent}}  & \textit{{\small target}}  & \textit{{\small sent}} \\
    \midrule
	{\small SS ~\cite{li2017sentimentscope}}& 46.3 & 36.9 & 44.6 & 33.4 & 20.2 & 14.5 \\
	{\small SS + \textit{emb}~\cite{li2017sentimentscope}}& 57.1  & 48.0  & 46.8  & 33.5  & 35.9  & 24.1 \\
	{\small SA-CRF} &  60.8 & 51.4  &  49.7 & 34.0 &  54.2 & 43.4 \\
	{\small \textbf{EI-}} & 57.7 & 48.2 & 47.2 & 33.7 & 52.8 & 38.9\\
	{\small \textbf{EI}} & {\bf 62.0} & {\bf 51.6} & {\bf 50.0} & {\bf 34.2} & {\bf 54.4} & {\bf 43.4} \\
	\bottomrule
\end{tabular}
}
\caption{\color{black} $F_1$ scores of targets (\textit{target}) and their associated sentiment (\textit{sent}) on SemEval 2016 Restaurant Dataset.}
\label{tab:additional}
\end{table}

\section{Conclusion and Future Work}
In this work, we argue that properly modeling both \textit{explicit} structures in the output space and the \textit{implicit} structures in the input space are crucial for building a successful targeted sentiment analysis system. Specifically, we propose a new model that captures explicit structures with latent CRF, and uses LSTM and self-attention to capture rich implicit structures in the input space efficiently. 
Through extensive experiments, we show that our model is able to outperform competitive baseline models significantly, thanks to its ability to properly capture both explicit and implicit structural information.

Future work includes exploring approaches to capture explicit and implicit structural information to other sentiment analysis tasks and other structured prediction problems.

\section*{Acknowledgments}
{\color{black}We would like to thank the anonymous reviewers for their thoughtful and constructive comments. 
This work is supported by Singapore Ministry of Education Academic Research Fund (AcRF) Tier 2 Project MOE2017-T2-1-156.}

\bibliography{emnlp-ijcnlp-2019}
\bibliographystyle{acl_natbib}

\appendix
\section{Appendix}
\label{sec:app}

\begin{figure}[h]
 \centering
 \begin{adjustwidth}{0.5cm}{0.0cm}
 \scalebox{1.0}
  {
  \includegraphics[width=\linewidth]{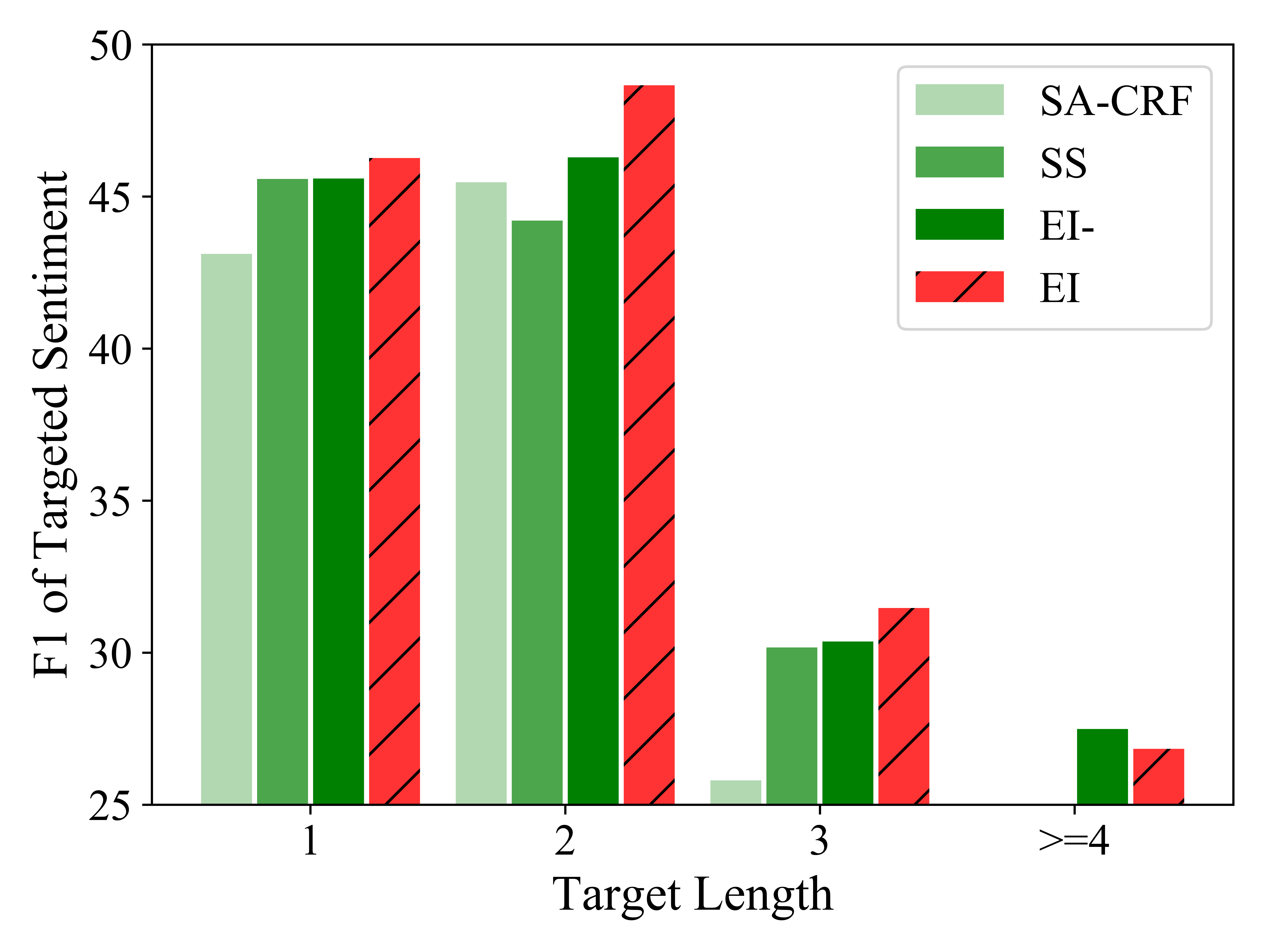}
  }
    \caption{Results of different lengths on English}
  \label{fig:chunklength_en}
  \end{adjustwidth}
\end{figure}

\begin{table}[h]
\centering

\begin{subtable}{.5\textwidth}
\centering
\begin{small}
\begin{tabular}{l c c c c c}
\toprule
 & \#instance & \#target &  \#$+$ & \#$-$ & \#$0$ \\
\midrule
Train & 1,925 &3,078 &  2384 & 475 & 219 \\
Test & 1,209 &  952 &  654 & 203 & 95 \\
\bottomrule
\end{tabular}
\end{small}
\vspace{-2mm}
\caption{Statistics on Russian.}
\vspace{1mm}
\end{subtable}

\begin{subtable}{.5\textwidth}
\centering
\begin{small}
\begin{tabular}{l c c c c c}
\toprule
 & \#instance & \#target &  \#$+$ & \#$-$ & \#$0$ \\
\midrule
Train & 674 &894 &  513 & 287 & 94 \\
Test & 575 &  373 &  229 & 120 & 24 \\
\bottomrule
\end{tabular}
\end{small}
\vspace{-2mm}
\caption{Statistics on Dutch.}
\vspace{1mm}
\end{subtable}

\begin{subtable}{.5\textwidth}
\centering
\begin{small}
\begin{tabular}{l c c c c c}
\toprule
 & \#instance & \#target &  \#$+$ & \#$-$ & \#$0$ \\
\midrule
Train & 1,234 &1,743 &  1,236 & 438 & 69 \\
Test & 676 &  612 & 468 & 114 & 30 \\
\bottomrule
\end{tabular}
\end{small}
\vspace{-2mm}
\caption{Statistics on English.}
\vspace{1mm}
\end{subtable}

\caption{Corpus statistics of SemEval 2016 Restaurant Dataset}
\label{fig:training_add_data_stats}
\end{table}

\subsection{Robustness}
We also report the results for targets of different lengths on English in Figure~\ref{fig:chunklength_en}.
As we can see, our model \textbf{BI} outperforms others except when the length is greater than or equal 4.
Note that according to statistics in the main paper, there exists a small number of targets of length 4.

\subsection{Additional Experiments}
We present the data statistics for English, Dutch and Russian in SemEval 2016 Restaurant dataset~\cite{pontiki2016semeval} in Table~\ref{fig:training_add_data_stats}.

\end{document}